\documentclass{ecai} 

\usepackage{latexsym}
\usepackage{amssymb}
\usepackage{amsmath}
\usepackage{amsthm}
\usepackage{booktabs}
\usepackage{enumitem}
\usepackage{graphicx}
\usepackage{color}
\usepackage{multirow}
\usepackage{algorithm}
\usepackage{algorithmic}

\usepackage{listings}

\newcommand{\BibTeX}{B\kern-.05em{\sc i\kern-.025em b}\kern-.08em\TeX}

\begin{document}


\begin{frontmatter}




\title{Learning Future Representation with Synthetic Observations for Sample-efficient Reinforcement Learning}


\author[A,B]{\fnms{Xin}~\snm{Liu}\thanks{Email: liuxin2021@ia.ac.cn.}}
\author[A,B]{\fnms{Yaran}~\snm{Chen}}
\author[A,B]{\fnms{Dongbin}~\snm{Zhao}} 

\address[A]{State Key Laboratory of Multimodal Artificial Intelligence Systems, \\
Institute of Automation, Chinese Academy of Sciences}
\address[B]{School of Artificial Intelligence, University of Chinese Academy of Sciences}

\begin{abstract}
    In visual Reinforcement Learning (RL), upstream representation learning largely determines the effect of downstream policy learning. Employing auxiliary tasks allows the agent to enhance visual representation in a targeted manner, thereby improving the sample efficiency and performance of downstream RL. Prior advanced auxiliary tasks all focus on how to extract as much information as possible from limited experience (including observations, actions, and rewards) through their different auxiliary objectives, whereas in this article, we first start from another perspective: auxiliary training data. We try to improve auxiliary representation learning for RL by enriching auxiliary training data, proposing \textbf{L}earning \textbf{F}uture representation with \textbf{S}ynthetic observations \textbf{(LFS)}, a novel self-supervised RL approach. Specifically, we propose a training-free method to synthesize observations that may contain future information, as well as a data selection approach to eliminate unqualified synthetic noise. The remaining synthetic observations and real observations then serve as the auxiliary data to achieve a clustering-based temporal association task for representation learning. LFS allows the agent to access and learn observations that have not yet appeared in advance, so as to quickly understand and exploit them when they occur later. In addition, LFS does not rely on rewards or actions, which means it has a wider scope of application (e.g., learning from video) than recent advanced auxiliary tasks. Extensive experiments demonstrate that our LFS exhibits state-of-the-art RL sample efficiency on challenging continuous control and enables advanced visual pre-training based on action-free video demonstrations. 
\end{abstract}

\end{frontmatter}


\section{Introduction}

Representation learning plays a critical role in Deep Reinforcement Learning (DRL). Conventional DRL algorithms depend on reward functions to simultaneously learn feature representation and control policy, yielding success across various domains \citep{atari-rl,continuous-rl,bnas,ppo,haoran}
. However, when dealing with complex high-dimensional inputs like visual observation, these traditional methods suffer from low sample efficiency~\citep{simple}, requiring a large number of interaction steps and wall-clock training time to obtain an effective policy. This is mainly attributed to the unqualified representation learning. In conventional DRL, the only supervision signal (reward) doesn't directly judge the representation ability of the visual encoder~\citep{atc}, which leads to the lack of visual supervision in vanilla DRL. For the above issue, one effective and popular solution is to introduce Self-Supervised Learning (SSL) into DRL. Through unsupervised auxiliary tasks, SSL is able to improve the neural perception module in a targeted manner. The labels of these tasks come from the training data itself, which means no additional manual annotation is required. SSL has achieved considerable results across various research fields like Computer Vision (CV)~\citep{cv1,simclr} and Natural Language Processing (NLP)~\citep{nlp2,nlp1}.

\begin{figure}[t]
    \centering
    \includegraphics[width=0.45\textwidth]{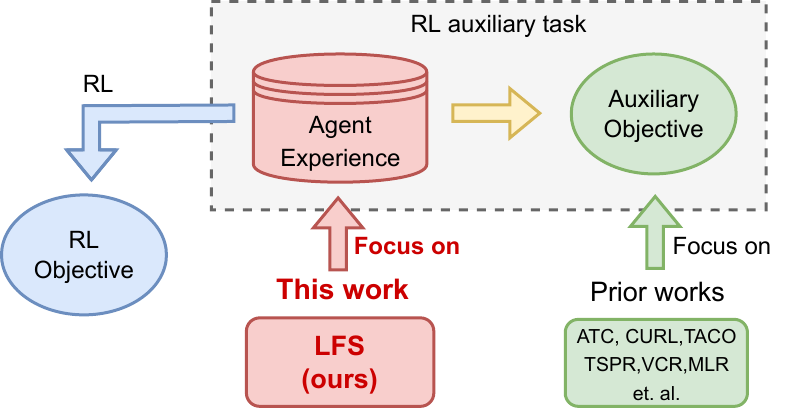}
    \caption{The difference between our proposed LFS and prior advanced auxiliary tasks. We show the general framework of RL auxiliary tasks, where the auxiliary objective is optimized over the data sampled from the collected experience. The data and objective together determine the auxiliary learning performance. Prior advanced methods design different auxiliary objectives to extract as much information related to decision making as possible from limited experience, while our LFS first starts from the perspective of enriching training data for auxiliary objectives.}
    \label{difference}
    \vspace{30pt}
\end{figure}

Inspired by their success, some works directly combine existing visual auxiliary objectives in CV with policy learning objectives, finding that the learned semantic representation can effectively improve the RL sample efficiency~\citep{sac-ae,cpc,curl}. The intervention of temporal information between neighboring observations further improves the representation learning performance in RL~\citep{atc}. 
In order to extract more representational information related to decision-making, subsequent methods attempt to design auxiliary objectives tailored for the Markov Decision Process (MDP) \citep{mdp}. They are no longer limited to observations, but involve rewards and actions in the auxiliary objective design~\citep{spr,efficientzero}. This is achieved by using transformers to encode action sequence information~\citep{mlr,minsong,mind}, training extra dynamic models for self-supervised labels~\citep{vcr,playvirtual}, or learning additional action representation modules~\citep{taco}. In summary, recent advanced approaches are committed to extracting as much information as possible from limited experience. However, complex auxiliary objective design and additional model training not only bring additional severe training burdens, but also limit the application scenarios of auxiliary tasks, e.g., value-free unsupervised RL and action-free visual pre-training. In addition, no matter how effective the extraction method (auxiliary objective) is, it is hard for these methods to break through the limitations caused by limited experience.

Instead of finding better objectives like prior works, we try to enhance the self-supervised RL by enriching training data. To the best of our knowledge, we are the first to start from the perspective of training data in auxiliary task design, which is shown in Figure \ref{difference}. The proposed approach, which we name \textbf{L}earning \textbf{F}uture representation with \textbf{S}ynthetic observations (\textbf{LFS}), aims to let agents access and learn the unseen future observations that have not yet appeared in experience, so as to quickly understand and exploit them when they later occur. First, we propose frame mask, a simple and training-free data synthesis method to produce observations that may contain future information. It draws inspiration from frame stack, a common visual RL preprocessing technique that concatenates consecutive historical frames to form observations as RL inputs. The proposed frame mask selectively concatenates nonconsecutive frames to synthesize novel observations that are currently unseen but may emerge in the future. A clear example to illustrate the motivation of our frame mask is provided in Figure \ref{framemask} left. Apparently, not all of the synthetic observations are qualified for representation learning. We correspondingly propose a data selection approach named Latent Nearest-neighbor Clip (LNC), which eliminates noisy synthetic observations based on real experience in the latent semantic space. The remaining synthetic observations and real observations together serve as the auxiliary training data, where we employ a clustering-based temporal association task for representation learning. Unlike recent advanced self-supervised auxiliary tasks, LFS does not require extra model training and does not rely on action or reward information. It enables effective visual representation learning that supports efficient policy learning in different scenarios, including end-to-end RL and action-free video pre-training.

We conduct extensive numerical and analytical experiments across different challenging continuous control tasks, including balance control, locomotion, and manipulation. The results demonstrate that our LFS outperforms state-of-the-art RL auxiliary tasks on sample efficiency. It even surpasses many advanced unsupervised RL pre-training methods on challenging few-shot manipulation tasks, without the need for a long reward-free pre-training phase. Moreover, we demonstrate that LFS can pre-train an effective and generic visual encoder over non-expert videos, which recent advanced auxiliary tasks cannot achieve due to their dependence on actions and rewards. We also provide detailed analytical experiments to validate our motivation. 

The contributions of our paper can be summarized as follows:
\begin{itemize}
    \item We first try to enhance the self-supervised RL from the perspective of enriching auxiliary training data, proposing Learning Future representation with Synthetic observations (LFS), a novel self-supervised auxiliary task for visual RL. LFS aims to let agents access and learn the observations that are currently unseen in experience, so as to quickly understand and exploit them when they occur later. 
    \item Based on the frame stack in visual RL, we propose frame mask to generate novel synthetic observations without extra training. To reduce the influence of unqualified synthetic data, we correspondingly design a data selection method LNC and employ a clustering-based temporal association task to learn visual representation without reward supervision.
    \item Extensive experiments on challenging continuous control tasks demonstrate that our LFS enables state-of-the-art sample efficiency in end-to-end RL. Without the pre-training process, it can outperform advanced unsupervised RL pre-training methods. Furthermore, we demonstrate that LFS is able to pre-train an effective and generic encoder over non-expert videos, which recent advanced auxiliary tasks cannot achieve. We also provide detailed analytical experiments to validate our motivation.
\end{itemize}

\section{Related Works}
\subsection{Sample Efficiency in RL}

Interacting with the environment to collect experience for RL is always a time-consuming and expensive process. In some special scenarios, such as autonomous driving \citep{driving} and real robot learning \citep{realrobot}, the interactions can even lead to danger. To this end, improving the sample and learning efficiency has always been one of the most concerning topics in the field of RL, especially visual RL. First, data augmentation, a CV technique to resist overfitting, was introduced into RL and achieved similar success in improving encoder robustness \citep{rad-sac,drq,drq-v2}. As a plug-and-play module, it now serves as a common technique combined with different types of sample-efficient RL methods. According to previous works \citep{minsong,curl,mind}, these methods can be categorized as follows: model-based RL, unsupervised RL, and self-supervised auxiliary tasks. Model-based RL methods \citep{simple,dreamer,dreamerv2,dreamerv3,dreamerpro} train extra world models based on RL experience. The world models in turn augment the RL experience, thereby improving sampling efficiency. Unsupervised RL \citep{comsd,aps,apt,diayn,protorl} assumes that accessing environments without extrinsic rewards is free. The related methods design intrinsic rewards to achieve policy or representation pre-training, reduce training burden in multi-task scenarios. Auxiliary tasks \citep{curl,taco,vcr,mlr,minsong} alleviate the lack of supervision signals for representation learning in DRL, accelerating downstream policy learning by improving the upstream representation understanding capabilities. This work falls into the last class, while we also utilize data augmentation in our design and are inspired by the idea of generating training data in model-based RL. Note that model-based methods synthesize experience for RL, while our LFS first proposes to synthesize observations for auxiliary objectives.

\begin{figure*}[t]
    \centering
    \includegraphics[width=0.98\textwidth]{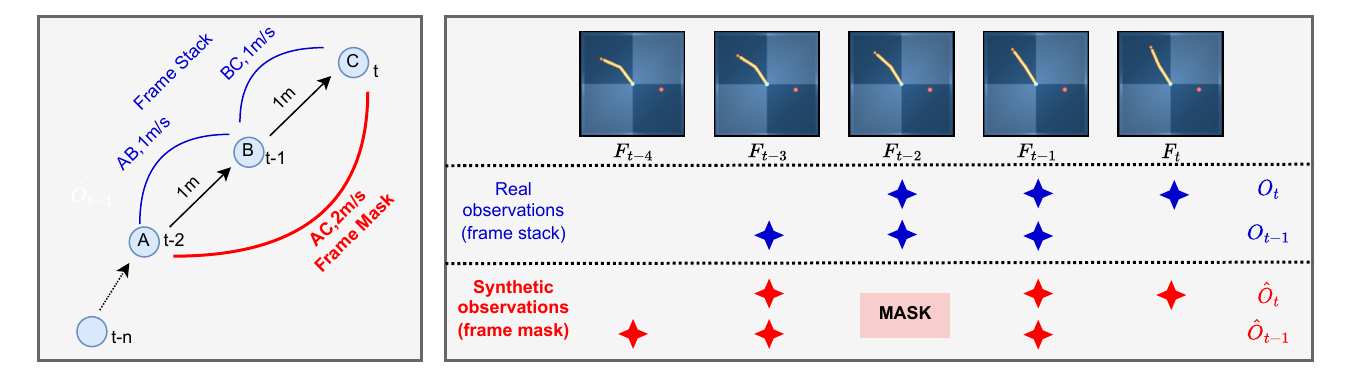}
    \caption{\textbf{Left}: A clear example to show why frame mask can generate novel observations that may contain future information. There's an agent on a platform, and the reward is positively related to agent speed. Its current speed is 1 m/s. A,B, and C are agent locations in three consecutive frames (1 second interval). The real observations (AB\&BC) generated by frame stack contain the speed information of 1 m/s. By masking frame B, frame mask produces a 'future' observation (AC) containing a speed of 2 m/s. \textbf{Right}: Difference between real observations and synthetic observations in our experiments, where the number of frame stack is set to three throughout the paper.}
    \label{framemask}
    \vspace{20pt}
\end{figure*}

\subsection{Self-supervised auxiliary tasks in RL}

Self-supervised learning is aimed at acquiring rich representation without the need for labels, achieving considerable results across different fields like CV \citep{simclr} and NLP \citep{nlp2}. There are also many auxiliary tasks that assist policy learning with rewards, such as future prediction \citep{earlysslrl}. For challenging visual RL, upstream representation learning greatly affects downstream policy learning. To this end, some works directly combined the existing visual self-supervised tasks in CV with DRL, finding that the improved visual representation can effectively help downstream policy learning in visual RL  \citep{cpc,curl}. With the temporal information added to the visual auxiliary tasks \citep{atc}, the policy performance is further improved. At the same time, it was found that the auxiliary tasks can independently support ideal representation learning without the help of the RL objective \citep{atc}, which inspires many works studying task-agnostic generic representation pre-training \citep{protorl,apt,crptpro,mvp}. To extract more information related to decision making, recent advanced works are not limited to observation, but utilize the actions or rewards in MDP to achieve auxiliary task design \citep{spr,efficientzero,vcr}. Some advanced works \citep{mlr,minsong,mind} employ Transformer to enhance the processing of sequential actions, while observation-action joint representation \citep{taco} also attracts much attention. These methods all strive to extract as much information from limited experience as possible, while our LFS is totally different, trying to enhance auxiliary representation learning by enriching training data. In addition, LFS doesn't require action or reward information, which means a more generalized application scenario.

\section{Methodology of LFS}

In this section, we introduce the details of our proposed LFS. In Section 3.1, we introduce how and why our proposed frame mask can synthesize unseen observations. To reduce the impact of unqualified synthetic observations, we correspondingly propose a selection method named Latent Nearest-neighbor Clip (LNC) to eliminate the noisy synthetic data, which we detail in Section 3.2. The selected synthetic observations are combined with real observations to form the auxiliary training data. We then achieve self-supervised representation learning through a clustering-based temporal association task, as illustrated in Section 3.3. Section 3.4 introduces how to utilize the proposed LFS. The pseudo code of LFS for end-to-end RL is available in Appendix A.

\subsection{Observation Synthesis via Frame Mask}


\paragraph{Frame stack.} First of all, we need to illustrate the frame stack, a common pre-processing technique for visual RL that inspires our frame mask. In most circumstances, a visual control task could be formulated as an infinite-horizon Partially Observable Markov Decision Process (POMDP) \cite{mdp}, denoted by $\mathcal{M}^p=(\mathcal{F},\mathcal{A},\mathcal{P},\mathcal{R},\gamma,d_0)$, where $\mathcal{F}$ is the high-dimensional image space, $\mathcal{A}$ is the action space, $\mathcal{P}$ is the distribution of next image given the history and current action, $R$ is the reward function, $\gamma$ is the discount factor, and $d_0$ is the distribution of the initial observation. In most circumstances, it's are to directly obtain ideal policy through RL over a POMDP because necessary temporal information related to sequential decision making, such as speed and movement, is missing in only one frame. To this end, frame stack, which concatenates several consecutive previous pixels as the current observation, is proposed and becomes a common pre-processing technique for RL. With frame stack, a POMDP is converted into a Markov Decision Process (MDP) $\mathcal{M}= (\mathcal{O},\mathcal{A},\mathcal{P}',\mathcal{R}',\gamma,d_0')$, where current observation contains temporal information and agents can make proper decisions based on it without exposure to history. 

\paragraph{Motivation of frame mask.} The proposed frame mask draws inspiration from frame stack. It produces synthetic observations by masking some historical frames and concatenating the rest. In other words, frame mask concatenates non-consecutive historical frames instead of consecutive frames. The time difference between two non-consecutive frames is larger than that between two consecutive frames, which means greater physical changes in some cases. To this end, the synthetic observations may contain greater physical changes that correspond to more advanced temporal information (e.g., higher speed or more active movement). These advanced observations cannot be encountered by the current policy but may be sampled by a better future policy. We use an example to clearly show our motivation, as shown in Figure \ref{framemask} left.

\paragraph{Concrete process of frame mask.} After demonstrating the motivation, we provide a detailed process for the proposed frame mask, as shown in Figure \ref{framemask} right. As per convention in visual RL, we set the number of stacking frames to three throughout this paper, which means one observation contains three frames. Given the current frame $F_{t}$ and four consecutive historical frames [$F_{t-1},F_{t-2},F_{t-3},F_{t-4}$], the real observations $O_{t}$ and its neighbor $O_{t-1}$ can be obtained by frame stack:

\begin{eqnarray}
\begin{aligned}
O_{t} &= \rm concat \it (F_{t},F_{t-1},F_{t-2})   \\
O_{t-1} &= \rm concat \it (F_{t-1},F_{t-2},F_{t-3}),
\end{aligned}
\end{eqnarray}

\noindent  where $\rm concat(\cdot) \it$ concatenates frames in the channel dimension. The proposed frame mask ignores (masks) the middle frame $F_{t-2}$ and sequentially stacks the rest frames, producing synthetic observation $\hat{O}^{t}_{t}$ and its neighbor $\hat{O}^{t}_{t-1}$: 

\begin{eqnarray}
\begin{aligned}
\hat{O}^{t}_{t} &= \rm concat \it (F_{t},F_{t-1},F_{t-3})   \\
\hat{O}^{t}_{t-1} &= \rm concat \it (F_{t-1},F_{t-3},F_{t-4}).
\end{aligned}
\end{eqnarray}

Note that $\hat{O}^{t}_{t-1}$ sampled at time $t$ and $\hat{O}^{t-1}_{t-1}$ sampled at time $t-1$ are not the same, where $\hat{O}^{t-1}_{t-1}$ is actually $\rm concat \it (F_{t-1},F_{t-2},F_{t-4})$. Then, $O_{t}$, $O_{t-1}$, action $a_{t-1}$, and reward $r_{t-1}$ are stored as a transition into the experience buffer $B_{rl}$ for RL training, which follows the standard paradigm of off-policy RL \citep{continuous-rl}. $(\hat{O}^{t}_{t},\hat{O}^{t}_{t-1})$ is saved into the auxiliary buffer $B_{a}$ without actions and rewards. $B_{rl}$ and $B_{a}$ provide data for auxiliary representation learning together, which we describe in the next two sections.

\subsection{Data Selection via Latent Nearest-neighbor Clip}

In Section 3.1, frame mask produces synthetic observations that may contain future information. However, not all the synthetic data is qualified for auxiliary representation learning. For example, some synthetic observations may violate kinematics, being out of distribution. To this end, we design a data selection approach, named Latent Nearest-neighbor Clip (LNC), aiming to eliminate those unqualified synthetic observations for robuster self-supervised learning. This is achieved by using real observations as a benchmark and selecting synthetic observations that are within a medium distance. The motivation is that the far synthetic observations are highly likely to be out of distribution, while the near synthetic observations are similar to real observations, providing limited novel information.

\begin{figure}[t]
    \centering
    \includegraphics[width=0.35\textwidth]{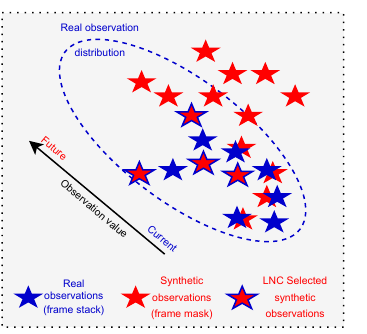}
    \caption{Latent Nearest-neighbor Clip (LNC) selects synthetic observations that are within a medium nearest-neighbor distance from real observations in the latent space. The motivation is that the far synthetic observations are highly likely to be out of distribution, while the near synthetic observations can't provide enough extra information. }
    \label{LNC}
    \vspace{30pt}
\end{figure}

Considering that (i) we don't know the latent distribution of real observations and (ii) high-dimensional distance calculations are expensive, we employ a particle-based k-nearest-neighbor Euclidean distance in the latent space as the metric, which is shown in Figure \ref{LNC}. Specifically, a batch of transitions is sampled from RL experience $B_{rl}$ for off-policy RL, where $M$ denotes the batch size. After data augmentation, these transitions are denoted as $\{(O_{t_i-1},a_{t_i-1},r_{t_i-1},O_{t_i} \}^{M}_{i=1}$. M synthetic observation pairs $\{(\hat{O}^{t_j}_{t_j},\hat{O}^{t_j}_{t_j-1})\}^{M}_{j=1}$ are sampled from auxiliary buffer $B_{a}$. The boundary of clip is based on the average k-nearest-neighbor Euclidean distance calculated over the sampled RL batch in the latent space:

\begin{eqnarray}
D = \frac{1}{M} \sum_{i = 1}^{M} ||h_{t_i-1}-h_{t_{k-nn(i)}-1}||,
\end{eqnarray}

\noindent where $h_{t_i-1}$ denotes the latent embedding of $O_{t_i-1}$ and $h_{t_{k-nn(i)}-1}$ denotes the latent embedding of $h_{t_i-1}$'s k-nearest neighbor in the RL batch. Clip center and clip range are calculated by $c\cdot D$ and $r\cdot D$ respectively, where $c$ and $r$ are fixed hyper-parameters. Correspondingly, the clip range is $[(c-r/2)D,(c+r/2)D]$.

For each synthetic observation pair $(\hat{O}^{t_j}_{t_j-1},\hat{O}^{t_j}_{t_j})$, we calculate its Euclidean distance from its k-nearest-neighbor real observation in the latent space:

\begin{eqnarray}
D_j = ||\hat{h}_{t_j-1}-h_{t_{k-nn(j)}-1}||,
\end{eqnarray}

\noindent where $\hat{h}_{t_j-1}$ denotes the latent embedding of $\hat{O}^{t_j}_{t_j-1}$ and $h_{t_{k-nn(j)}-1}$ denotes the latent embedding of $h_{t_j-1}$'s k-nearest neighbor in the RL batch. Now, the synthetic observation pairs whose $D_j$ are within the clip range $[(c-r/2)D,(c+r/2)D]$ are selected. They are defined as $\{(\hat{O}^{t_j}_{t_j},\hat{O}^{t_j}_{t_j-1})\}^{N}_{j=1}$, where $N$ denotes the number of LNC-selected synthetic observation pairs. In addition, $M-N$ real observation pairs with data augmentation are randomly selected from the sampled batch. These $M-N$ real observation pairs and $N$ selected synthetic observation pairs together serve as the auxiliary training data, collectively defined as $\{(\bar{O}^{(x)}_{t-1},\bar{O}^{(x)}_{t})\}^{M}_{x=1}$.

\begin{figure*}[t]
    \centering
    \includegraphics[width=0.85\textwidth]{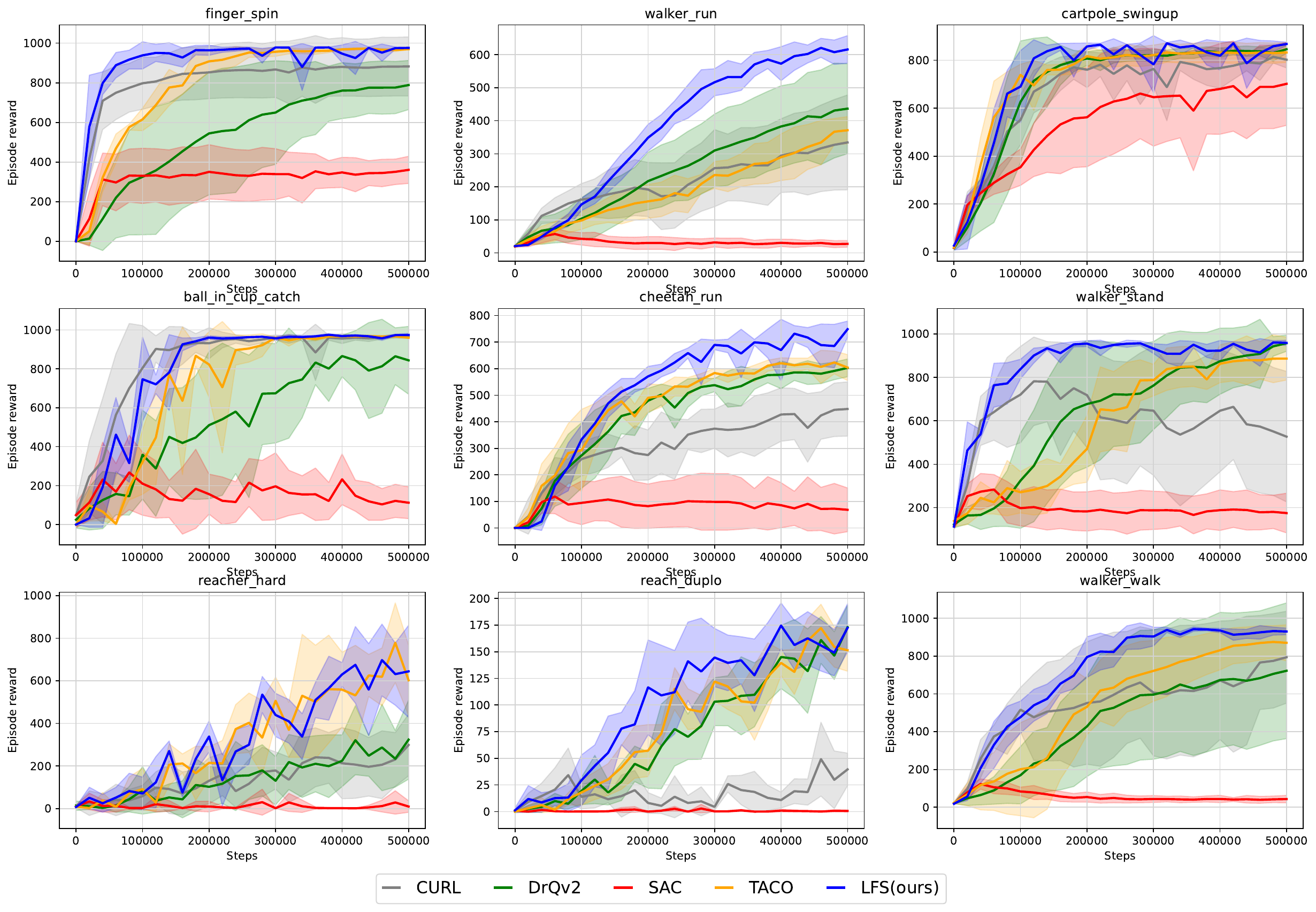}
    
    \caption{The training curve of LFS and four competitive end-to-end RL baselines on nine challenging continuous control tasks. LFS is the only self-supervised RL method that significantly outperforms DrQv2 and SAC across all nine tasks, overall exhibiting state-of-the-art RL sample efficiency. It proves that our main idea, enhancing auxiliary representation learning by enriching auxiliary data, is of feasibility and rationality.}
    \label{dmcontrol-endtoend}
    \vspace{20pt}
\end{figure*}

\subsection{Clustering-based Temporal Association}

With the $\{(\bar{O}^{(x)}_{t-1},\bar{O}^{(x)}_{t})\}^{M}_{x=1}$ as auxiliary data, self-supervised representation learning can be achieved by optimizing the auxiliary objective. Although prior advanced works design different auxiliary objectives, we find that they all aim to enhance the understanding of movement by adding temporal information. To this end, we employ a temporal association task, which requires the model to associate one observation with its neighbor. Currently, contrastive learning, which aims to optimize infoNCE loss \citep{cpc}, is the most popular way to achieve association tasks. While in our LFS, we employ a clustering-based objective instead of infoNCE loss, where the unsupervised label of each observation is computed over the whole training batch instead of the single association target. According to previous studies \citep{dreamerpro,swav,protorl}, clustering-based objectives lead to a more stable and efficient training process for small batch sizes with unstable distribution. Motivated by this, we utilize it to alleviate the impact of the unstable data distribution brought by synthetic observations, achieving more stable representation and policy learning in experiments. 

The objective of LFS is calculated as follows: First, each last observation $\bar{O}^{(x)}_{t-1}$ is encoded by the convolutional visual encoder $f_\theta$, projected by a Multi-Layer Perceptron (MLP) $g_\theta$, and predicted by another MLP $v_\psi$ to produce $z^{(x)}_{t-1}$. These two MLPs don't change the dimension of the latent embedding. Then, we take a softmax over the dot product between $z^{(x)}_{t-1}$ and M trainable prototypes $\{c_y\}_{y=1}^M$ that serve as the clustering centers:
\begin{equation}
p^{(x)}_{t-1} = \rm{softmax} \it \left(\frac{\dot{z}_{t-1}^{(x)} \cdot \dot{c}_1}{\tau},...,\frac{\dot{z}_{t-1}^{(x)} \cdot \dot{c}_M}{\tau}\right),
\end{equation}

\noindent where ${\tau}$ denotes a temperature hyper-parameter and the dot over $z$ and $c$ is the $l_2$-normalization. For each current observation $\bar{O}^{(x)}_{t}$, it undergoes encoding by target $f_{\Bar{\theta}}$ and projection by target $g_{\Bar{\theta}}$ in turn to produce $\{z^{(x)}_{t}\}_{x=1}^M$ analogously. In order to avoid trival solutions \citep{protorl}, these target network are updated through the Exponential Moving Average (EMA) \citep{ema} instead of backpropagation, which is formulated as $\Bar{\theta} \xleftarrow{} (1-\eta)\Bar{\theta} + \eta\theta.$ Then Sinkhorn-Knopp clustering \cite{sinkhorn,simclr,protorl} algorithm is employed on $\{z^{(x)}_{t}\}_{x=1}^M$ and prototypes $\{c_y\}_{y=1}^M$ to obtain the batch-clustering assignment targets $\{q^{(x)}_{t}\}_{x=1}^M$ for all training observations. We refer readers to \citet{simclr} for the concrete process of Sinkhorn-Knopp clustering. The objective is:

\begin{equation}
    \mathcal{J}_{LFS} = -\frac{1}{M}\sum_{x=1}^M q^{{(x)}_{t}^T}\rm log \it p^{(x)}_{t-1}.
\end{equation}

When optimizing $\mathcal{J}_{LFS}$, $\theta$, $\psi$, and $\{c_y\}_{y=1}^M$ are updated. The learned Encoder $f_\theta$ is shared in visual RL.

\subsection{Utilizing LFS}
Since LFS does not rely on actions or rewards, it has a wider range of application scenarios. First, it enables sample-efficient end-to-end RL and we employ SAC \citep{sac} as the backbone RL algorithm. When optimizing the SAC objective, we detach the visual encoder from the computational graph, decoupling representation learning from policy learning. This reduces the training burden without hurting learning performance \citep{atc}. We provide the pseudo code of end-to-end LFS in Appendix A. Moreover, LFS can be utilized for pre-training on video demonstrations. This is achieved by optimizing LFS objective over the sequential observations directly obtained from videos. LFS enables generic visual pre-training on non-expert video demonstrations from multiple domains. Without the need for finetuning, the pre-trained encoder can be implemented for efficient downstream RL on both seen and unseen tasks.

\section{Experiments \& Analysis}

\subsection{End-to-end RL on Continuous Control}

In this section, we show the end-to-end RL comparison on challenging continuous control tasks from DeepMind Control suite (DMControl). We first compare our LFS with four baselines: TACO \citep{taco}, CURL \citep{curl}, DrQv2 \citep{drq-v2}, and SAC \citep{sac}. Among them, TACO is a state-of-the-art self-supervised RL approach, DrQv2 is a state-of-the-art naive RL method (only optimizing the RL objective), and SAC is the backbone RL algorithm of the proposed LFS. CURL is also a popular auxiliary task for RL. These methods are comprehensively compared on nine challenging and representative continuous control tasks, including robot locomotion (\textit{walker run}), manipulation (\textit{reach duplo}), and balance control (\textit{finger spin}). On each task, each method is run for 500K steps and evaluated for 10 episodes every 20K steps. The results of LFS are averaged over five random seeds. Hyper-parameters are provided in Appendix B.

\begin{table*}[t]
\caption{We further compare end-to-end LFS with unsupervised RL methods on four manipulation tasks defined in the URLB benchmark. The eight unsupervised RL methods require large numbers of reward-free steps (1M steps here) for policy pre-training and another few steps with task rewards (100k steps here) for policy finetuning. Without the need for extra pre-training, our LFS can learn effective policies within only 100k steps, surpassing most pre-training baselines. In comparison, end-to-end expert DrQv2 cannot outperform four of them, e.g., ICM.}
\vspace{20pt}
\centering
\renewcommand\arraystretch{1.1}
\begin{tabular}{|c|cc|cccccccc|}
\hline
                                                        & \multicolumn{2}{c|}{No pre-training}                                                                       & \multicolumn{8}{c|}{Pre-training with intrinsic reward for 1M steps}                                                                                                                                                                                                                                                                                                                                                                          \\
\multirow{-2}{*}{Task}                                  & \begin{tabular}[c]{@{}c@{}}LFS\\ (ours)\end{tabular} & \begin{tabular}[c]{@{}c@{}}DrQv2\\ \citep{drq-v2}\end{tabular} & \begin{tabular}[c]{@{}c@{}}ICM\\ \citep{icm}\end{tabular} & \begin{tabular}[c]{@{}c@{}}Disagreement\\ \citep{disagrement}\end{tabular} & \begin{tabular}[c]{@{}c@{}}RND\\ \citep{rnd}\end{tabular} & \begin{tabular}[c]{@{}c@{}}APT\\ \citep{apt}\end{tabular} & \begin{tabular}[c]{@{}c@{}}Proto-RL\\ \citep{protorl}\end{tabular} & \begin{tabular}[c]{@{}c@{}}SMM\\ \citep{smm}\end{tabular} & \begin{tabular}[c]{@{}c@{}}DIAYN\\ \citep{diayn}\end{tabular} & \begin{tabular}[c]{@{}c@{}}APS\\ \citep{aps}\end{tabular} \\ \hline
{ \textit{jaco reach bottom left}}  & { 59±12}                         & { 23±10}                        & { \textbf{65±19}}             & { 42±10}                               & { 46±9}                       & 0±0                                               & 36±13                                                  & 1±1                                               & 8±3                                                 & 0±0                                               \\
{ \textit{jaco reach bottom right}} & { \textbf{142±46}}               & { 23±8}                         & { 88±23}                      & { 58±11}                               & { 44±9}                       & 0±0                                               & 43±10                                                  & 1±0                                               & 4±1                                                 & 0±0                                               \\
{ \textit{jaco reach top left}}     & { \textbf{154±47}}               & { 40±9}                         & { 76±19}                      & { 89±19}                               & { 59±7}                       & 6±4                                               & 41±10                                                  & 2±1                                               & 20±4                                                & 2±0                                               \\
{ \textit{jaco reach top right}}    & { \textbf{113±66}}               & { 37±9}                         & { 87±24}                      & { 49±12}                               & { 47±7}                       & 2±1                                               & 47±12                                                  & 4±3                                               & 22±6                                                & 5±1                                               \\ \hline
Mean score                                              & \textbf{117}                                         & 31                                                  & 79                                                & 60                                                         & 49                                                & 2                                                 & 42                                                     & 2                                                 & 14                                                  & 2                                                 \\ \hline
\end{tabular}
\label{urlb}
\end{table*}

\begin{table*}[t]

\caption{The comparison between three action-free auxiliary tasks on video pre-training. Two advanced unsupervised active visual pre-training methods are also included in the comparison. For each method, all nine downstream tasks share the same pre-trained encoder except end-to-end DrQv2, which serves as an expert to show the score level of different tasks. LFS enables better downstream policy performance than all other pre-training methods.}
\vspace{20pt}
\centering
\renewcommand\arraystretch{1.1}
\setlength{\tabcolsep}{3mm}{
\begin{tabular}{|c|ccccc|c|}
\hline
                                                                                       & \multicolumn{3}{c|}{Pre-training on videos}                                                                                                                                        & \multicolumn{2}{c|}{Pre-training with URL}                                                                 & \multicolumn{1}{l|}{End-to-end Expert}              \\
\multirow{-2}{*}{\begin{tabular}[c]{@{}c@{}}Task\\ (sharing one encoder)\end{tabular}} & \begin{tabular}[c]{@{}c@{}}LFS\\ (ours)\end{tabular} & \begin{tabular}[c]{@{}c@{}}ATC\\ \citep{atc}\end{tabular} & \multicolumn{1}{c|}{\begin{tabular}[c]{@{}c@{}}CURL\\ \citep{curl}\end{tabular}} & \begin{tabular}[c]{@{}c@{}}Proto-RL\\ \citep{protorl}\end{tabular} & \begin{tabular}[c]{@{}c@{}}APT\\ \citep{apt}\end{tabular} & \begin{tabular}[c]{@{}c@{}}DrQv2\\ \citep{drq-v2}\end{tabular} \\ \hline
{ \textit{reach duplo}}                                            & { \textbf{126±20}}               & { 2±2}                        & { 10±9}                                             & { 2±3}                             & { 1±0}                        & 173±21                                              \\
{ \textit{cartpole swingup}}                                       & { \textbf{736±104}}              & { 428±174}                    & { 586±70}                                           & { 562±67}                          & { 668±73}                     & 846±17                                              \\
{ \textit{walker run}}                                             & { \textbf{524±28}}               & { 196±12}                     & { 191±8}                                            & { 367±11}                          & { 187±3}                      & 437±136                                             \\
{ \textit{reacher hard}}                                           & { \textbf{129±104}}              & { 45±63}                      & { 112±83}                                           & { 19±34}                           & { 56±40}                      & 324±186                                             \\
{ \textit{cheetah run}}                                            & { \textbf{586±26}}               & { 292±10}                     & { 193±13}                                           & { 298±26}                          & { 310±13}                     & 603±33                                              \\
{ \textit{walker stand}}                                           & { \textbf{941±32}}               & { \textbf{933±12}}            & { \textbf{939±16}}                                  & { 553±372}                         & { 915±14}                     & 957±37                                              \\
{ \textit{ball in cup catch}}                                   & { \textbf{966±10}}               & { 888±29}                     & { \textbf{952±18}}                                  & { \textbf{949±7}}                  & { 932±3}                      & 844±174                                             \\
{ \textit{walker walk}}                                            & { \textbf{944±20}}               & { 805±37}                     & { 583±13}                                           & { 899±19}                          & { 696±20}                     & 722±359                                             \\
{ \textit{finger spin}}                                            & { \textbf{977±4}}                & { 917±6}                      & { 561±8}                                            & { 801±175}                         & { 467±3}                      & 789±124                                             \\ \hline
Mean score                                                                            & \textbf{659}                                         & 501                                               & 459                                                                     & 494                                                    & 470                                               & \textbf{-}                                          \\
Mean expert-normalized score                                                          & \textbf{0.982}                                       & 0.653                                             & 0.612                                                                   & 0.670                                                  & 0.614                                             & -                                                   \\ \hline
\end{tabular}}
\label{pre-training}
\end{table*}

The training curves are shown in Figure \ref{dmcontrol-endtoend}. LFS achieves better final policy performance and RL sample efficiency than CURL on 8/9 tasks, except \textit{ball in cup catch}. Compared with TACO, LFS is competitive on \textit{reacher hard}, \textit{reach duplo}, and \textit{cartpole swingup} and enables better sample efficiency on the other six tasks. In addition, LFS significantly outperforms SAC and DrQv2 across all nine tasks, which CURL and TACO cannot achieve. In summary, the results demonstrate that LFS overall enables better sample efficiency than prior state-of-the-art self-supervised methods, which proves
that our main motivation, enhancing self-supervised RL by expanding auxiliary data, is feasible and meaningful. In addition to above four baselines, we further make comparisons with another two advanced RL auxiliary tasks: VCR \citep{vcr} and MLR \citep{mlr} in Appendix B, where we obtain conclusions similar to those in this section. Due to non-open source or huge computational consumption, we utilize their paper-reported numerical scores. We also provide extra comparisons with model-based Dreamer \citep{dreamer} and DreamerV3 \citep{dreamerv3} in Appendix B.

\subsection{Comparison with Unsupervised RL}

To further demonstrate the sample efficiency advantages of the proposed LFS, we compare it with unsupervised RL methods, which employ a long task-agnostic pre-training stage before few-shot task-specific policy learning to improve sample efficiency. 
We are interested in whether LFS can achieve effective policy learning within only a few steps without unsupervised pre-training. We make comparisons with eight advanced and popular unsupervised RL methods on four manipulation tasks defined in the URLB benchmark \citep{urlb}. The eight methods are ICM \citep{icm}, Disagreement \citep{disagrement}, RND \citep{rnd}, APT \citep{apt}, Proto-RL \citep{protorl}, SMM \citep{smm}, DIAYN \citep{diayn}, and APS \citep{aps}. 1M environment steps without task rewards and 100K environment steps with task rewards are allowed for each unsupervised RL method. For our LFS and DrQv2, only 100K steps with task rewards are allowed for end-to-end RL. The hyper-parameter settings are shown in Appendix C.

The results are shown in Table \ref{urlb}. First, we observe that after 100k steps, end-to-end DrQv2 begins to understand the environment but has not yet learned effective strategies. Second, some unsupervised RL methods don't learn anything, even after a long unsupervised pre-training phase. The above phenomena fully demonstrate the difficulty of these four tasks. In this situation, LFS still obtains qualified policies on 3/4 tasks within only 100k steps, significantly outperforming all other baselines by a large margin on these three tasks. Numerically, LFS gets an average score of 117 on four tasks, which outperforms the next best ICM by $1.48\times$. Note that LFS doesn't have pre-training, while ICM does. DrQv2, which also has no pre-training stage, only gets 31. In summary, the results further confirm the superiority of end-to-end LFS over RL sample efficiency.

\begin{figure*}[t]
    \centering
    \includegraphics[width=0.95\textwidth]{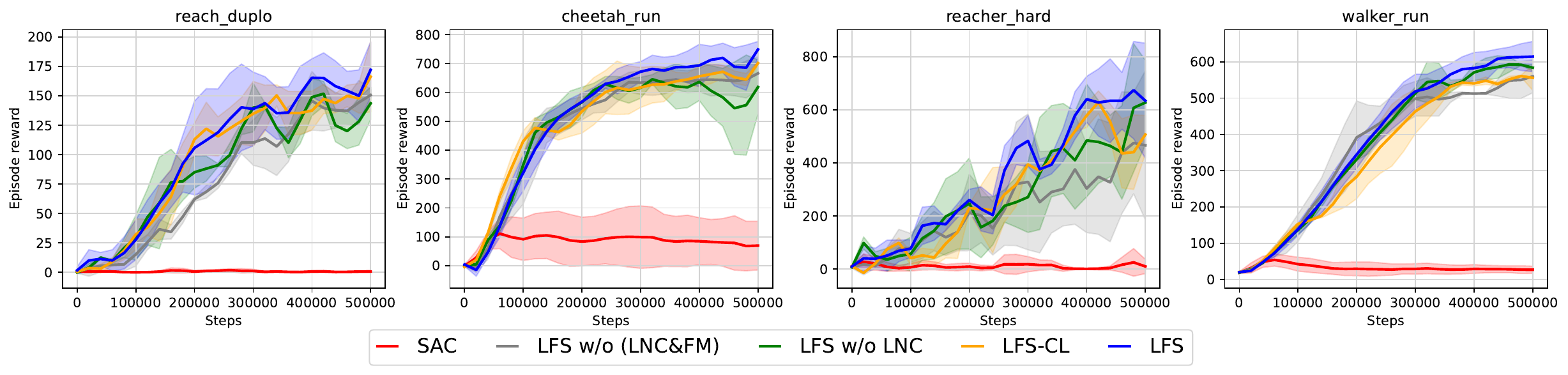}
    \caption{The ablation study of LFS. We separately ablate the proposed data selection method LNC (LFS w/o LNC), remove all synthetic observations produced by proposed frame mask (LFS w/o LNC\&FM), and utilize contrastive learning instead of clustering-based objective (LFS-CL), to show their respective effects on RL performance. Our backbone RL algorithm, SAC, is also shown in the figure. Curves are smoothed by Savitzky-Golay filter.}
    \label{ablation1}
    \vspace{30pt}
\end{figure*}

\begin{figure}[]
    \centering
    \includegraphics[width=0.48\textwidth]{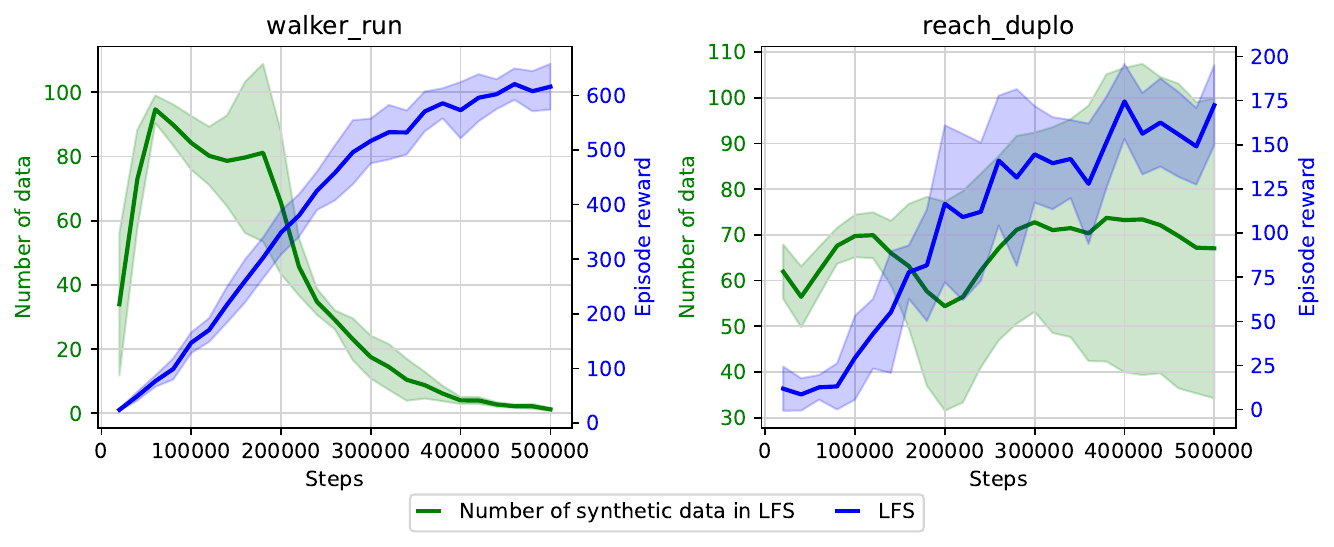}
    \caption{The number of selected observations in LFS training.}
    \label{LNC-number}
    \vspace{30pt}
\end{figure}

\subsection{Pre-training Generic Representation on Videos}

Different from most recent advanced methods (e.g., TACO \citep{taco}, MLR \citep{mlr}, and VCR \citep{vcr}), our LFS doesn't rely on actions or rewards to compute auxiliary objectives, having more application scenarios where actions are absent. In this section, we demonstrate that our LFS can pre-train effective and generic representation over non-expert video demonstrations. Concretely, we provide non-expert videos of three tasks to LFS: \textit{reach duplo}, \textit{walker run}, and \textit{cartpole swingup}. For each task, 320 non-expert demonstrations generated by a uniform random policy are provided, and each has 250 frames. We pre-train one encoder on demonstrations of all three tasks simultaneously with LFS for a total of 60K update times. The two action-free baselines, CURL \citep{curl} and ATC \citep{atc}, share the same experimental settings as LFS. In addition, we further employ two state-of-the-art unsupervised active pre-training methods in comparison. They are allowed to explore the reward-free environments for 240K steps of unsupervised pre-training (equal to 60K encoder update times). After pre-training, each pre-trained encoder is frozen and used to conduct 500K steps of downstream RL on both seen and unseen tasks. It means all nine tasks share the same representation module, which is really challenging. DrQv2 \citep{drq-v2} (500K steps on each task) serves as a end-to-end expert to show the score level of different tasks.  

The results shown in Table \ref{pre-training} demonstrate that the encoder pre-trained with LFS objective leads across all nine tasks, beating both the video pre-training baselines and unsupervised active pre-training baselines. Concretely, LFS obtains an average score of 659, outperforming the next best video pre-training method, ATC by $1.32\times$. This is mainly attributed to the LFS's novel ability to enrich self-supervised training data. In comparison, active pre-training methods are allowed to explore the environment for training data collection, but they are still not as good as our method. This is due to their unstable pre-training process, where they need to train data collection agents and learn representation simultaneously. By contrast, LFS can enrich limited experience without extra model training. Compared with expert DrQv2, LFS leads on 4/9 tasks and achieves an 0.982 average normalized score, which means they have similar performance. Note that in this section, each pre-trained encoder is frozen and shared across all nine tasks, while it is not required for DrQv2. In conclusion, prior advanced methods are either unable to work in an action-free scenario, or inferior to LFS in pre-training performance.

\subsection{Ablation Study}
In this section, we ablate different parts of our LFS to show their influence, which is shown in Figure \ref{ablation1}. First, we ablate LNC, employing a fixed number of randomly selected synthetic observations, which is denoted as LFS w/o LNC. We find that without LNC, LFS is unstable and often exhibits performance drops (e.g., in \textit{cheetah run}) when agents have adapted to the environment and learned effective policy. This verifies that there is much noise in synthetic observations, which can be effectively alleviated by LNC. Second, we remove all the synthetic observations sampled by frame stack (FM), which we denote as LFS w/o (LNC\&FM) in the figure. The performance also degenerates, which demonstrates that our selected synthetic data contains helpful novel information that assists and improves auxiliary representation learning. In summary, both the synthetic observations of frame mask and LNC are necessary in our LFS. In addition, as shown in Section 3.3, the LFS objective can also be implemented by contrastive learning (CL), which is denoted as LFS-CL in the figure. Although LFS-CL also enables sample-efficient RL across all the tasks, the comparison confirms that employing clustering-based objectives is a better choice when (i) the batch size $M$ is small and (ii) the distribution of training data is not stable enough.

\subsection{Effect of LNC}
In LFS, we propose LNC to determine which synthetic observations are utilized as auxiliary data. In Figure \ref{LNC-number}, we visualize the number of selected synthetic observations in the end-to-end RL process. The selection number decreases as the agent strategy improves in locomotion (\textit{walker run}) while being less affected in manipulation (\textit{reach duplo}). This is consistent with intuition. In locomotion, movement posture is the key. The more well-trained the locomotion strategy, the stronger the correlation between neighboring frames, and the greater the impact of masking frames on movement semantics. As training proceeds, the noise generated by frame mask will become more and more, and the proposed LNC effectively eliminate it. In contrast, the movement of a robotic arm is much simpler in manipulation. Masking frames don't drastically change the semantics in most cases. Therefore, frame mask is more stable for manipulation than locomotion, leading to more stable LNC selection process.

\begin{table}[t]
\centering
\renewcommand\arraystretch{1.1}

\caption{Value comparison of synthetic and real observations. }
\vspace{20pt}
\setlength{\tabcolsep}{0.8mm}{
\begin{tabular}{|c|cc|c|cc|}
\hline
{ \textit{reach duplo}} & { mean value}           & { max value}            & { \textit{cheetah run}} & { mean value}          & max value           \\ \hline
{ synt obs}             & { \textbf{148.4±1.1}} & { \textbf{219.1±0.7}} & { synt obs}             & { \textbf{83.9±0.6}} & \textbf{94.4±0.5} \\
{ real obs}             & { 142.2±0.7}         & { 216.3±0.8}          & { real obs}             & { 80.9±0.7}          & 92.5±0.7          \\ \hline
\end{tabular}}
\label{value}
\end{table}

\subsection{Does LFS Access Future Observations?}
In this section, we examine whether LFS can access unseen observations in advance by analyzing observation values. In the rapid rise stage of LFS training (200K steps), we test the values of selected synthetic observations and real observations. We employ the learned RL critic model with a uniform policy to compute the values of each observation in each update. The results are averaged over the next 1K updates, as shown in Table \ref{value}. They demonstrate that the synthetic observations exhibit higher scores on both value metrics. Note that synthetic observations are not utilized in RL critic model training, which means they have natural disadvantages on these two critic-based metrics. In conclusion, LFS indeed accesses the observations whose values are bigger than the real observations sampled by current policy, which means it indeed achieves self-supervised learning with the 'future' observations.

\section{Conclusion \& Limitation}
In this paper, we propose LFS, a novel self-supervised auxiliary task for visual RL, to improve its sample efficiency, with extensive experiments to verify and analyze its effectiveness. 
To the best of our knowledge, LFS is the first self-supervised RL algorithm that focuses on enriching auxiliary training data, which is totally different from prior advanced methods that are dedicated to designing better auxiliary objectives. We hope this novel idea (enhancing self-supervised RL via enriching auxiliary data) can inspire researchers, so that future related works are no longer limited to auxiliary objective design. There are also some limitations that can be improved. For example, the proposed frame mask is a relatively naive method. In domains where temporal information (e.g., speed and movement) is not critical or action space is discrete, it is hard to generate helpful observations. In addition, frame mask is also powerless to generate novel actions. Inspired by the success of the diffusion model, employing generative models to enrich self-supervised data is a potential solution, which we leave to future study.







\bibliography{ecai-sample-and-instructions}

\appendix

\onecolumn

\clearpage
\section{Pseudo code}

We provide the pseudo-code of LFS for end-to-end RL, which is shown in Algorithm \ref{algorithmlfs}. 
\begin{algorithm}[]
\caption{Pseudo-code of LFS.}
\textsc{\#\#\#} Utilizing LFS for end-to-end RL.

\textbf{Require:} Reward-specific task environment $E$, environment training steps $I_{total}$, initial random exploration steps $I_{init}$, environment action repeat $a_{repeat}$.

\textbf{Initialize:} The visual encoder $f_{\theta}$, the actor model $A$ (the policy), the critic model $Q$, RL replay buffer $B_{rl}$, the auxiliary replay buffer $B_a$, and a queue $F_Q$ whose capacity is 5 to temporarily save frames.

\begin{algorithmic}[1]

\STATE \hspace{0cm}\textbf{for} $t = 1,...,I_{init}/a_{repeat}$ \textbf{do}
\STATE \hspace{0.5cm} Given current real observation $O_t$, sample action $a_t$ via a random uniform policy.
\STATE \hspace{0.5cm} Employ $a_t$ to interact $a_{repeat}$ times with $E$, obtaining reward $r_t$ and next real observation $O_{t+1}$.
\STATE \hspace{0.5cm} Add the transition $(O_t,a_t,r_t,O_{t+1})$ into RL replay buffer $B_{rl}$.
\STATE \hspace{0.5cm} Obtain current frames $F_t$ from current observation $O_t$.
\STATE \hspace{0.5cm} Save the current frame $F_t$ into the queue $F_Q$. 
\STATE \hspace{0.5cm}\textbf{if} queue $F_Q$ is full \textbf{do}
\STATE \hspace{1cm} Get the $(F_t,F_{t-2},F_{t-3},F_{t-4},F_{t-5})$ out of the $F_Q$.
\STATE \hspace{1cm} Produce synthetic observation pairs $(\hat{O}^{t}_{t},\hat{O}^{t}_{t-1})$ via Equation (2).
\STATE \hspace{1cm} Save $(\hat{O}^{t}_{t},\hat{O}^{t}_{t-1})$ into $B_a$.

\STATE \hspace{0cm}\textbf{for} $t = I_{init}/a_{repeat}+1,...,I_{total}/a_{repeat} $ \textbf{do}
\STATE \hspace{0.5cm} Given the current real observation $O_t$, encode it with $f_{\theta}$.
\STATE \hspace{0.5cm} Sample action $a_t$ by actor $A$ according to the latent image embedding.
\STATE \hspace{0.5cm} Employ $a_t$ to interact $a_{repeat}$ times with $E$, obtaining reward $r_t$ and next real observation $O_{t+1}$.
\STATE \hspace{0.5cm} Add the transition $(O_t,a_t,r_t,O_{t+1})$ into RL replay buffer $B_{rl}$.
\STATE \hspace{0.5cm} Obtain current frame $F_t$ from the current observation $O_t$.
\STATE \hspace{0.5cm} Save the current frame $F_t$ into queue $F_Q$. 
\STATE \hspace{0.5cm}\textbf{if} queue $F_Q$ is full \textbf{do}
\STATE \hspace{1cm} Get the $(F_t,F_{t-2},F_{t-3},F_{t-4},F_{t-5})$ out of the $F_Q$.
\STATE \hspace{1cm} Produce synthetic observation pairs $(\hat{O}^{t}_{t},\hat{O}^{t}_{t-1})$ via Equation (2).
\STATE \hspace{1cm} Save $(\hat{O}^{t}_{t},\hat{O}^{t}_{t-1})$ into $B_a$. 

\STATE \hspace{0.5cm} Sample M real transition $\{(O_{t_i-1},a_{t_i-1},r_{t_i-1},O_{t_i} \}^{M}_{i=1}$ from $B_{rl}$ with data augmentation.
\STATE \hspace{0.5cm} Sample M synthetic observation pairs $\{(\hat{O}^{t_j}_{t_j},\hat{O}^{t_j}_{t_j-1})\}^{M}_{j=1}$, from $B_{a}$.
\STATE \hspace{0.5cm} Compute the internal distance $D$ via equation (3) and obtain clip range $[(c-r/2)D,(c+r/2)D]$.
\STATE \hspace{0.5cm} Compute the distance from synthetic data to real data $D_j$ via equation (4).
\STATE \hspace{0.5cm} Select $\{(\hat{O}^{t_j}_{t_j},\hat{O}^{t_j}_{t_j-1})\}^{N}_{j=1}$ whose $D_j$ is within $[(c-r/2)D,(c+r/2)D]$ and select $M-N$ real observation pairs randomly.
\STATE \hspace{0.5cm} Combine them together as $\{(\bar{O}^{(x)}_{t-1},\bar{O}^{(x)}_{t})\}^{M}_{x=1}$.
\STATE \hspace{0.5cm} Using $\{(\bar{O}^{(x)}_{t-1},\bar{O}^{(x)}_{t})\}^{M}_{x=1}$ to compute the auxiliary objective via Equation (6) and update the encoder $f_{\theta}$.

\STATE \hspace{0.5cm} Detach the encoder $f_{\theta}$ and use $\{(O_{t_i-1},a_{t_i-1},r_{t_i-1},O_{t_i} \}^{M}_{i=1}$ to update actor $A$ and critic $Q$ via SAC.

\STATE \hspace{0cm}\textbf{end for}
\end{algorithmic}
\textbf{Output:} The well-trained policy $A$.

\label{algorithmlfs}
\end{algorithm}

\clearpage

We also provide the Pytorch-style implementions of both the proposed frame mask and LNC, which is shown in Algorithm \ref{pytorchframemask} and Algorithm \ref{pytorchlnc} respectively.

\begin{algorithm}[H]

\caption{PyTorch-style pseudo code for frame mask.}
\scriptsize
\begin{lstlisting}[language=Python]
define self.obs_quene to save the historical frames by reserving real observations.

def frame mask(self):
        assert (self.obs_quene[0][6:,:,:] == self.obs_quene[2][:3,:,:]).all()
        assert (self.obs_quene[2][6:,:,:] == self.obs_quene[4][:3,:,:]).all()
        assert (self.obs_quene[2][6:,:,:] == self.obs_quene[3][3:6,:,:]).all()

        # extract the frames from real observations and mask the middle frame.
        frame1 = self.obs_quene[2][:3,:,:]
        frame2 = self.obs_quene[2][3:6,:,:]
        frame4 = self.obs_quene[4][3:6,:,:]
        frame5 = self.obs_quene[4][6:,:,:]

        # synthesize a novel observation and its neighbor via Equation (2).
        syn_obs = np.concatenate([frame1,frame2,frame4],axis=0)
        syn_next_obs = np.concatenate([frame2,frame4,frame5],axis=0)
        assert syn_obs.shape == (9,84,84)
        
        return syn_obs, syn_next_obs
\end{lstlisting}
\label{pytorchframemask}
\end{algorithm}

\begin{algorithm}[H]

\caption{PyTorch-style pseudo code for LNC.}
\scriptsize
\begin{lstlisting}[language=Python]
def LNC(self, obs_synthetic, obs_original):

        # obtaining latent embeddings
        with torch.no_grad():
            obs_synthetic = self.encoder(obs_synthetic)
            obs_original = self.encoder(obs_original)

        # compute k-nearest-neighbor distance from synthetic data to real data via Equation (4)
        define k.
        all_d = torch.norm(obs_synthetic[:, None, :] - obs_original[None, :, :], dim=2, p=2)
        d, _ = torch.topk(all_d, k, dim=1, largest=False)
        distance = torch.reshape(d,(-1,))
        assert distance.size()==torch.Size([obs_synthetic.shape[0]])
        assert distance.size(0)==obs_synthetic.shape[0]
        
        # compute the internal k-nearest-neighbor distance between real observations via Equation (3)
        internal_d = torch.norm(obs_original[:, None, :] - obs_original[None, :, :], dim=2, p=2)
        d_internal_0self, _ = torch.topk(internal_d, k+1, dim=1, largest=False)
        assert d_internal_0self[0,0].item()==0
        assert d_internal_0self[1,0].item()==0
        mean_d_internal = torch.mean(d_internal_0self[:,-1:]).item()

        # compute the clip range
        define c (clip center coefficient).
        define r (clip range coefficient).
        clip_low = mean_d_internal*(c-r/2)
        clip_high = mean_d_internal*(c+r/2)

        # select the data within the clip range
        chosen_idxs = []
        for index in range(0,obs_synthetic.size(0)):
            if distance[index] > clip_low and distance[index] < clip_high:
                chosen_idxs.append(index)

        return chosen_idxs, len(chosen_idxs)
\end{lstlisting}
\label{pytorchlnc}
\end{algorithm}

\clearpage
\section{Additional Results}
\subsection{Additional end-to-end RL results}

As shown in Section 4.1, we further compare our proposed LFS with another two state-of-the-art RL auxiliary tasks: VCR \citep{vcr} and MLR \citep{mlr}. VCR doesn't open-source their code. MLR is available, but its computational consumption is too large due to its transformer-based auxiliary objective. On a NVIDIA Tesla V100 GPU, MLR requires more than 90 hours wall-clock time for 500K steps RL on \textit{walker walk}, while our LFS requires only 7.1 hours. Since our computing resources are limited, we list their paper-reported results here and don't draw their curves. In addition, we further compare our LFS with two other state-of-the-art model-based RL algorithms: Dreamer \citep{dreamer} and DreamerV3 \citep{dreamerv3}. The results shown in Table \ref{additionalendtoend} further demonstrate that our LFS enables state-of-the-art RL sample efficiency on challenging continuous control.

\begin{table}[]
\caption{The additional comparison between LFS and two state-of-the-art self-supervised RL algorithms on six continuous control tasks. Two advanced model-based RL methods are also compared in this table. For each task, 500K steps are allowed. LFS overall exhibits better policy learning performance. }
\vspace{20pt}
\centering
\renewcommand\arraystretch{1.25}
\setlength{\tabcolsep}{3mm}{
\begin{tabular}{|c|ccccc|}
\hline
                                                  & \multicolumn{3}{c|}{Self-supervised RL}                                                                                                                                           & \multicolumn{2}{c|}{Model-based RL}                                                                             \\
\multirow{-2}{*}{Task}                            & \begin{tabular}[c]{@{}c@{}}LFS\\ (ours)\end{tabular} & \begin{tabular}[c]{@{}c@{}}VCR\\ \citep{vcr}\end{tabular} & \multicolumn{1}{c|}{\begin{tabular}[c]{@{}c@{}}MLR\\ \citep{mlr}\end{tabular}} & \begin{tabular}[c]{@{}c@{}}Dreamer\\ \citep{dreamer}\end{tabular} & \begin{tabular}[c]{@{}c@{}}DreamerV3\\ \citep{dreamerv3}\end{tabular} \\ \hline
{ \textit{cheetah run}}       & { \textbf{749±30}}               & { 661±32}                     & { 674±37}                                          & { 570±253}                        & { 628±30}                           \\
{ \textit{cartpole swingup}}  & { \textbf{868±8}}                & { 854±26}                     & { \textbf{872±5}}                                  & { 762±27}                         & { 845±40}                           \\
{ \textit{walker run}}        & { \textbf{616±42}}               & { -}                          & { 576±25}                                          & { 460±97}                         & \textbf{613±117}                          \\
{ \textit{ball in cup catch}} & { \textbf{974±8}}                & { 958±4}                      & { \textbf{964±14}}                                 & { 879±87}                          & { \textbf{967±12}}                  \\
{ \textit{walker walk}}       & { \textbf{930±18}}               & { \textbf{930±18}}            & { \textbf{939±10}}                                 & { 897±49}                         & { \textbf{931±13}}                           \\
{ \textit{finger spin}}       & { \textbf{976±3}}                & { \textbf{972±25}}            & { \textbf{973±31}}                                 & { 796±183}                        & { 785±209}                          \\ \hline
\end{tabular}}
\label{additionalendtoend}
\end{table}

\subsection{Additional Ablations}

In addition to Section 4.4, we provide more ablation results in this section. The analysis is similar to that in Section 4.4.

\begin{figure*}[h]
    \centering
    \includegraphics[width=0.8\textwidth]{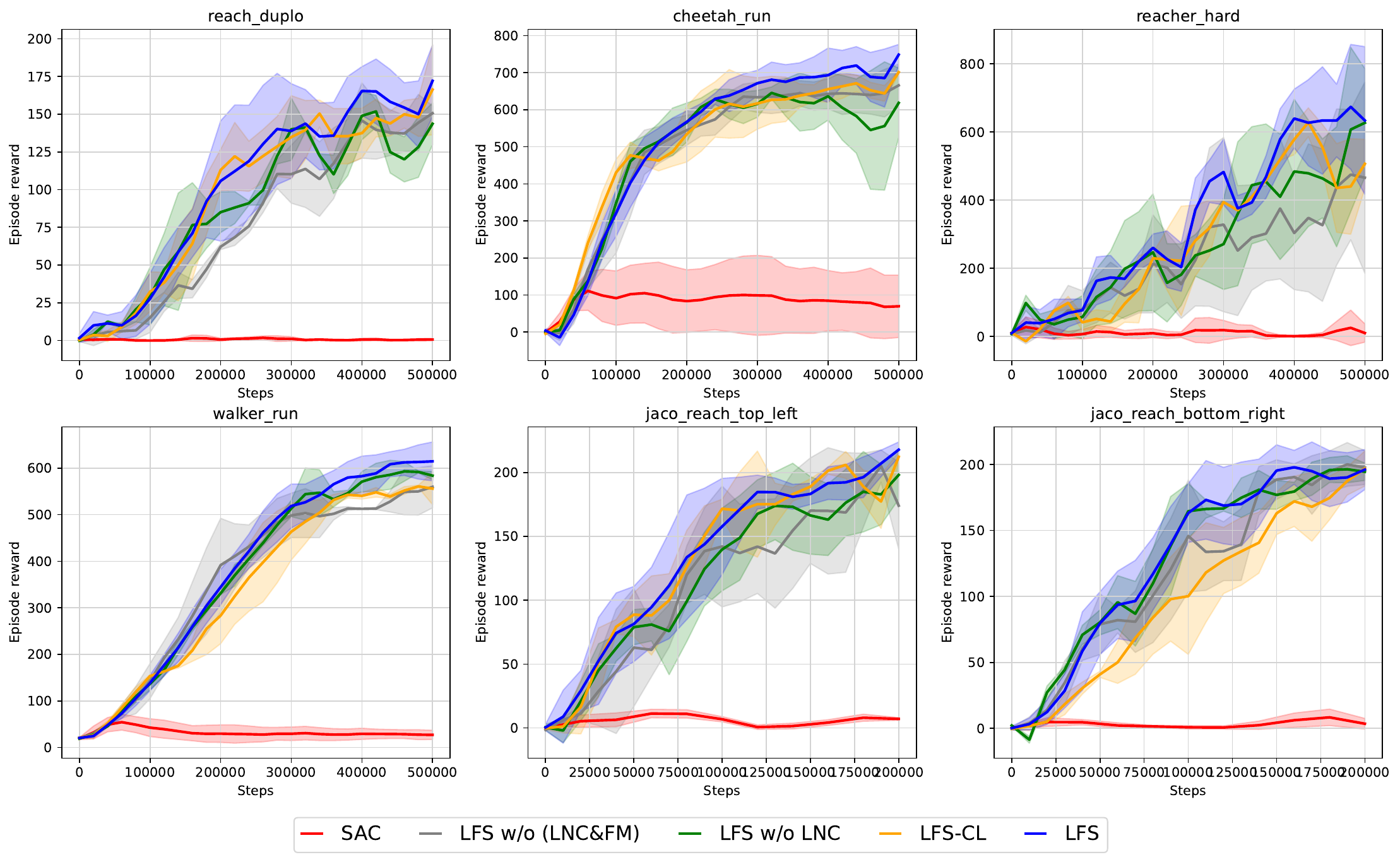}
    \caption{More ablation results of LFS. We separately ablate the proposed data selection method LNC (LFS w/o LNC), remove all synthetic observations produced by the proposed frame mask (LFS w/o LNC\&FM), and utilize contrastive learning instead of clustering-based objective (LFS-CL), to show their respective effects on RL performance. Our backbone RL algorithm, SAC, is also shown in the figure. In summary, they are all necessary designs in LFS. The curves are smoothed by the Savitzky-Golay filter.}
    \label{ablationall}
\end{figure*}

\clearpage
\subsection{Additional LNC Analysis}

In addition to Section 4.5, we provide more analytical results on LNC in this section. The two left works are locomotion tasks, while the two right works are manipulation tasks. We can obtain a similar conclusion to that in Section 4.5.

\begin{figure}[]
    \centering
    \includegraphics[width=0.6\textwidth]{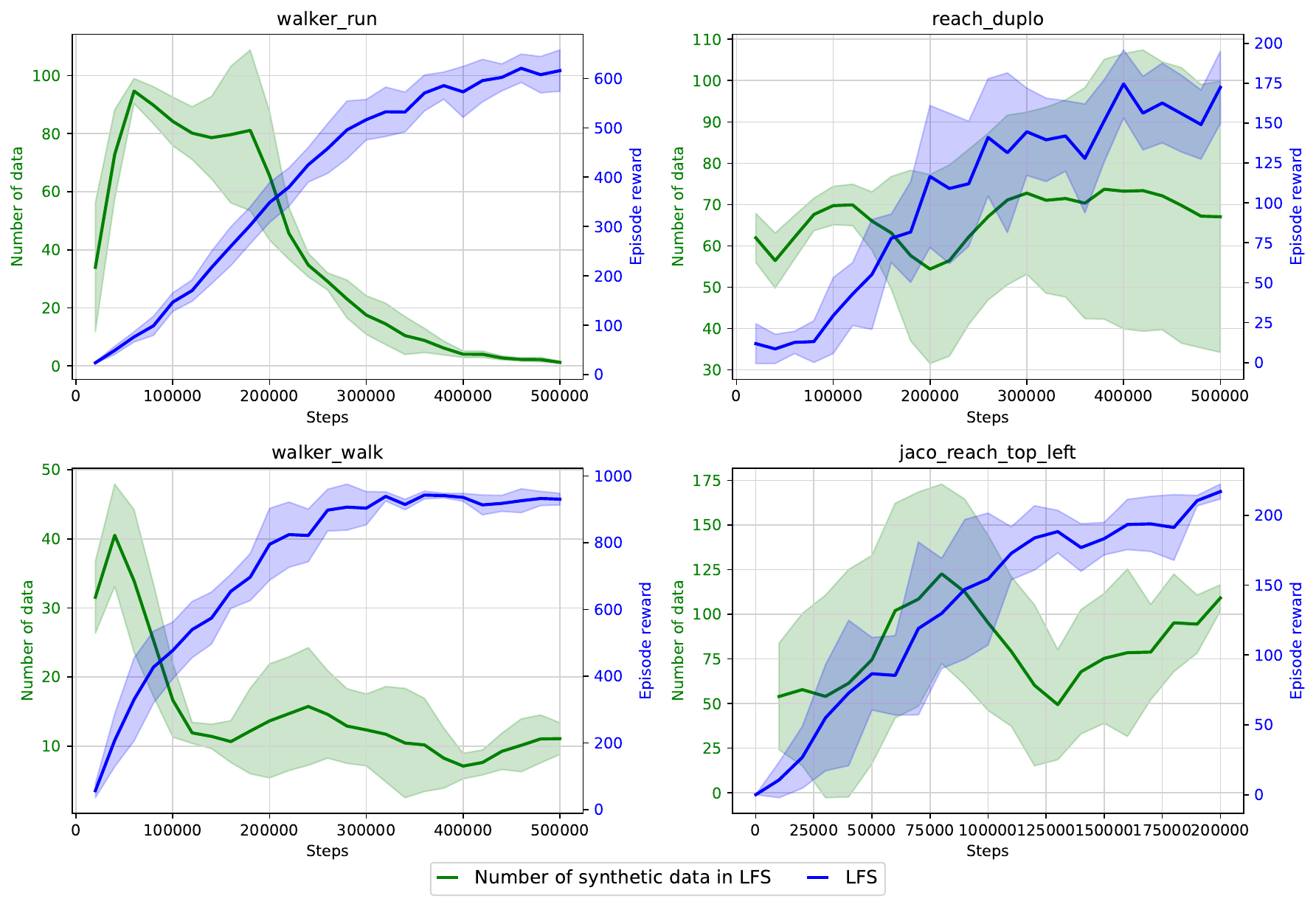}
    \caption{The additional result of analytical experiments on LNC.}
    \label{LNC-number}
\end{figure}

\clearpage
\section{Hyper-parameters}

We provide the detailed hyper-parameter settings of the proposed LFS and environments in this section. 

First, the settings of DMControl and URLB manipulation are shown in Table \ref{hyper-env}.


\begin{table}
\centering
\renewcommand\arraystretch{1.25}
\setlength{\tabcolsep}{3mm}{
\begin{tabular}{|l|c|}
\hline
Parameter                      & Setting       \\ \hline
Frame rendering          & $84\times84\times3$ RGB     \\
Frame stack number                   & $3$             \\
Observation dimensionality                   & $84\times84\times9$            \\
Action repeat                  & $2$             \\ \hline
\end{tabular}}
\vspace{10pt}
\caption{Settings of environments (DMControl and URLB manipulation) in our experiments.}
\label{hyper-env}
\end{table}

The default hyper-parameter settings for LFS are shown in \ref{hyper}. For DMControl tasks in Section 4.1, all the hyper-parameters follow the default settings, except LNC range coefficient $r$ set to 0.2 for \textit{walker run} and \textit{cheetah run}, and learning rate set to $1e-3$ for \textit{finger spin}. For URLB manipulation tasks in Section 4.2, LNC range coefficient $r$ set to 0.2. For video pre-training in Section 4.3, LNC center coefficient $c$ set to 0.6. In ablation, the number of synthetic observations in LFS w/o LNC is set to 52 ($0.1\cdot M$) according to the mean LNC number shown in Section 4.5. 

\begin{table}
\centering
\setlength{\tabcolsep}{3mm}{
\renewcommand\arraystretch{1.25}
\begin{tabular}{|l|c|}
\hline
Parameter                      & Setting       \\ \hline
Convolution channels             & $(32,32,32,32)$ \\
Convolution stride             & $(2,1,1,1)$     \\
Filter size                    & $3\times3$           \\
Representation dimensionality  & $39200$         \\
Latent dimensionality          & $128$           \\
Predictor hidden units         & $1024$          \\
Actor feature dimensionality   & $50$            \\
Actor MLP hidden units         & $1024$          \\
Critic feature dimensionality  & $50$            \\
Critic MLP hidden units        & $1024$          \\
Optimizer                   & Adam          \\
Learning rate                  & $1e-4$         \\
Number of clustering centers $M$         & $512$           \\
Batch size          & $512$           \\
Data augmentation    & Random shift   \\
Random shift pad               & $\pm4$             \\
Random shift bilinear interpolation & False  \\
Auxiliary buffer capacity                  & $40000$          \\
LNC $k$     &   $1$     \\
LNC center coefficient $c$     &   $0.9$     \\
LNC range coefficient $r$     &   $0.1$     \\
Encoder target update frequency     & $1$         \\
Encoder target EMA momentum     & $0.05$         \\
Backbone RL algorithm                   & SAC       \\
SAC initial temperature        & $0.1$           \\
Discount $\gamma$                      & $0.99$          \\
SAC replay buffer capacity                       & $40000$          \\
Actor update frequency         & $2$             \\
Actor log stddev bounds        & $[-10,2]$   \\
Critic update frequency        & $1$             \\
Critic target update frequency & $2$             \\
Critic target EMA momentum     & $0.01$          \\
Softmax temperature            & $0.1$           \\ \hline
\end{tabular}}
\vspace{10pt}
\caption{Default Hyper-parameter settings in LFS.}
\label{hyper}
\end{table}

\clearpage


\end{document}